\documentclass{llncs/llncs}

\usepackage[T1]{fontenc}  
\usepackage{graphicx} 
\usepackage{subfigure}

\def\review{false}

\def\reviewfalse{false}

\ifx\review\reviewfalse
\newcommand{\rtslam}{RT-SLAM}
\else
\newcommand{\rtslam}{\textit{X}}
\fi

\date{}

\title{\rtslam: a generic and real-time visual SLAM implementation}

\ifx\review\reviewfalse
\author{
Cyril Roussillon\inst{1,2,3} \and
Aurélien Gonzalez\inst{1,2} \and\\
Joan Solà\inst{1,2,4} \and
Jean-Marie Codol\inst{1,2,5} \and
Nicolas Mansard\inst{1,2} \and\\
Simon Lacroix\inst{1,2} \and
Michel Devy\inst{1,2}\\
~\\
{\texttt{\{firstname.name\}@laas.fr}}
}

\institute{
CNRS; LAAS; 7 avenue du colonel Roche; F-31077 Toulouse, France
\and
Université de Toulouse; UPS, INSA, INP, ISAE; LAAS; F-31077 Toulouse, France
\and
Funded by the Direction Générale de l'Armement (DGA), France
\and
Ictineu Submarins, Industria 12, 08980 St. Feliu de Llobregat, Barcelona, Catalonia
\and
NAV ON TIME, 42 avenue du Général De Croutte, 31100 Toulouse, France
}
\fi

\begin{document}

\maketitle

\begin{abstract}

  This article presents a new open-source C++ implementation to solve the SLAM
  problem, which is focused on genericity, versatility and high execution
  speed. It is based on an original object oriented architecture, that allows
  the combination of numerous sensors and landmark types, and the integration of
  various approaches proposed in the literature. The system capacities are
  illustrated by the presentation of an inertial/vision SLAM approach, for which
  several improvements over existing methods have been introduced, and that
  copes with very high dynamic motions. Results with a hand-held camera are
  presented.

\end{abstract}

\section{Motivation}

\setcounter{footnote}{0}

Progresses in image processing, the growth of available computing power and the
proposition of approaches to deal with bearings-only observations have made
visual SLAM very popular, particularly since the demonstration of a real-time
implementation by Davison in 2003 \cite{Dav_03}. The Extended Kalman Filter
(EKF) is widely used to solve the SLAM estimation problem, but it has recently been
challenged by global optimization methods that have shown superior precision for
large scale SLAM. Yet EKF still has the advantage of simplicity and faster
processing for problems of limited size \cite{Why_Filter_10}, and can be
combined with global optimization methods \cite{estrada05} to take the best of
both worlds.

Monocular EKF-SLAM reached maturity in 2006 with solutions for ini\-tia\-li\-zing
landmarks \cite{Kwok_04} \cite{Sola_05} \cite{Montiel_06}. Various methods for
landmark parametrization have been analyzed \cite{sola_ahp}, and the literature now
abounds with contributions to the problem.

This article presents \rtslam\footnote{\rtslam\ stands for
  \ifx\review\reviewfalse ``Real-Time SLAM'' \fi}, a software framework aimed at
fulfilling two essential requirements. The first one is the need for a generic,
efficient and flexible development tool, that allows to easily develop, evaluate
and test various approaches or improvements. The second one is the need for
practical solutions for live experiments on robots, for which localization and
mapping require real-time execution and robustness. \rtslam\ is a C++
implementation based on Extended Kalman Filter (EKF) that allows to easily
change robot, sensor and landmark models. It runs up to 60\,Hz with $640\times
480$ images, withstanding highly dynamic motions, which is required for instance
on humanoid or high speed all terrain robots. \rtslam\ is available as
\emph{open source} software at \ifx\review\reviewfalse
\texttt{http://rtslam.openrobots.org}.  \else \texttt{http:// }. 
\fi

Section \ref{sec-arch} details the architecture of \rtslam\ and section
\ref{sec-technique} presents some of the techniques currently integrated within,
to define an efficient inertial/visual SLAM approach. Section \ref{sec-results}
analyzes results obtained with a hand-held system assessed thanks to ground
truth, and section \ref{sec-conclusion} concludes the article by a presentation
of prospects.

\section{Overall Architecture\label{sec-arch}}

\begin{figure}[htb]
  \centering \subfigure[Main objects in a SLAM context. Different robots
  \texttt{Rob} have several sensors \texttt{Sen}, providing observations
  \texttt{Obs} of landmarks \texttt{Lmk}.  States of robots, sensors and
  landmarks are stored in the stochastic map \texttt{Map}. There is one
  observation per sensor-landmark pair.]{
\includegraphics[width=5.2cm]{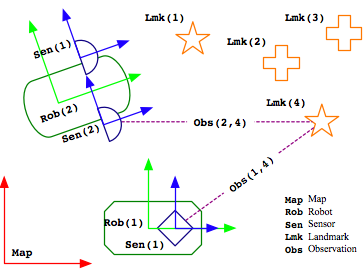}
\label{fig-rtslam-objects}
}
\hspace{0.2cm}
\subfigure[Objects hierarchy in \rtslam. Each individual map
\texttt{M} in the world \texttt{W} contains
robots \texttt{R} and landmarks \texttt{L}.
 A robot has sensors \texttt{S}, and an
observation \texttt{O} is created for every pair of sensor and landmark.
In order to allow full genericity, map managers \texttt{MM} and data managers
\texttt{DM} are introduced.]%
{
\includegraphics[width=6.1cm]{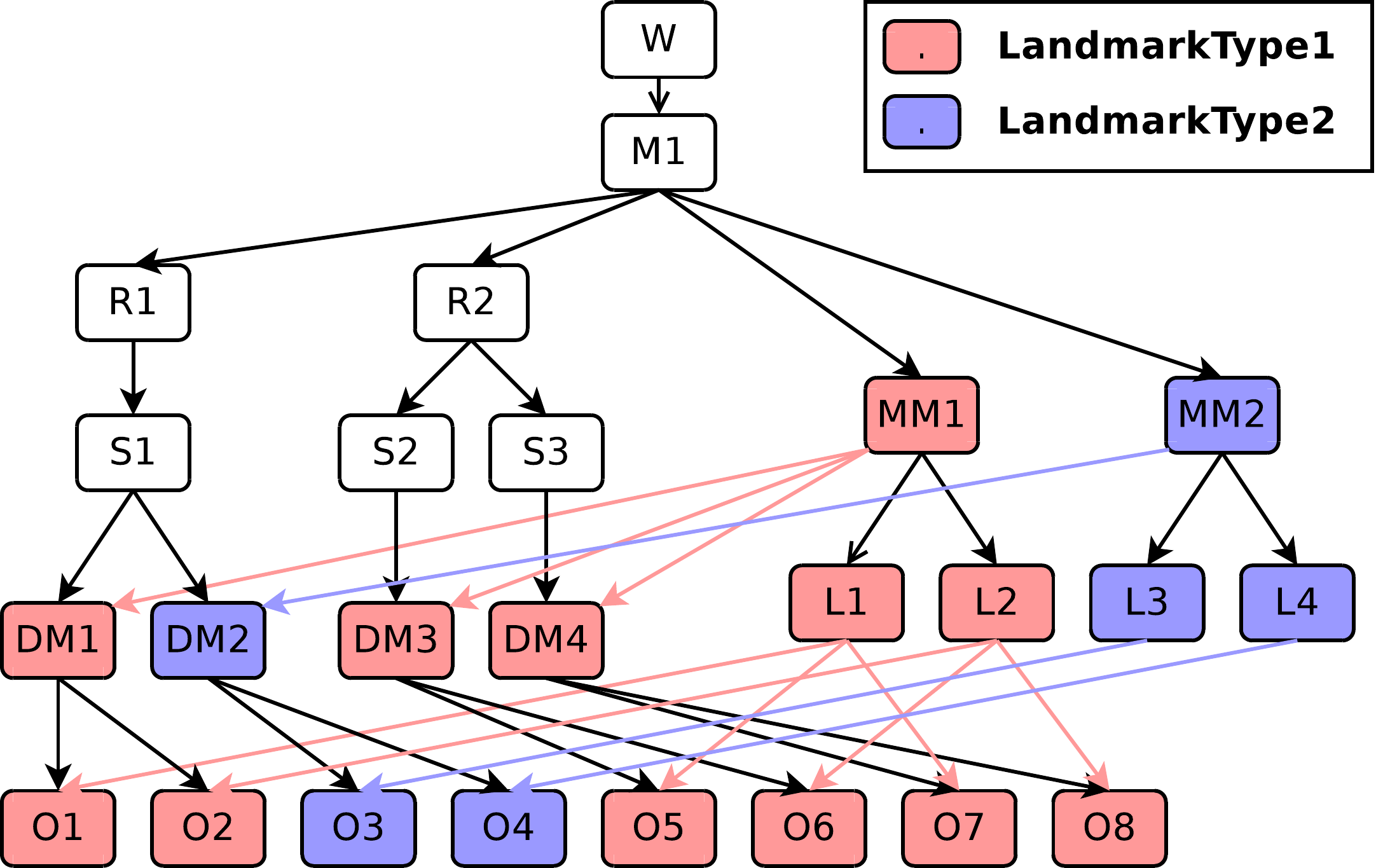}
\label{familyTree}
}
\caption{The main objects in \rtslam}
\label{rtslam-architecture}
\end{figure}

Fig. \ref{rtslam-architecture} presents the main objects defined in \rtslam.
They encompass the basic concepts of a SLAM solution: the \emph{world} or
environment contains \emph{maps}; maps contain \emph{robots} and
\emph{landmarks}; robots have \emph{sensors}; sensors make \emph{observations}
of landmarks. Each of these objects is abstract and can have different
implementations. They can also contain other objects that may themselves be
generic.

\paragraph{Map.}
The maps contain an optimization or estimation engine: for now
\rtslam\ uses a standard formulation of EKF-SLAM. Since this solution is very 
well documented in the
literature \cite{Dav_07}, it is not detailed in depth here. Indirect indexing
within Boost's ublas C++ structures is intensively
 used to exploit the sparsity of the problem and the symmetry of the
 covariances matrices.

\paragraph{Robot.}
Robots can be of different type according to the way their state is represented
and their prediction model. The latter can be either a simple kinematic model
(constant velocity, constant acceleration, \ldots) or a proprioceptive sensor
(odometric, inertial, \ldots), as illustrated section 
\ref{sec-prediction}. The proprioceptive sensor is an example of generic object
contained in robot objects, as different hardware can provide the same function.

\paragraph{Sensor.}
Similarly to robots, sensors can also have different models
(perspective camera, panoramic catadioptric camera, laser, \ldots), and
contain a generic exteroceptive sensor hardware object (firewire camera,
USB camera, \ldots).
In addition, as sensors belong to the map, their state can be estimated: this
opens the possibility for estimating other parameters such as extrinsic
calibration, time delays, biases and gain errors, and the like.

\paragraph{Landmark.}
Landmarks can be of different type (points, lines, planes, \ldots), and each
type can have different state parametrization (Euclidean point, inverse depth
point, \ldots).  Moreover the parametrization of a landmark can change over
time, as explained section \ref{sec-param}. A landmark also contains a
descriptor used for data association, which is a dual description to the state
representation.

As shown Fig. \ref{fig-rtslam-objects}, it is worth noticing that landmark
objects are common to the different sensors, all of them being able to observe
the same landmark (provided they have compatible descriptors for this landmark
of course).  This allows to greatly improve the observability of landmarks
compared to a system where the sensors are strictly independent. In the
particular case of two cameras for instance, landmarks can be used even if they
are only visible from one camera or if they are too far away for a stereovision
process to observe their depth (this process was introduced in \cite{sola_bicam}
as \emph{BiCam} SLAM).

\paragraph{Observation.}
In \rtslam, the notion of observation
plays a predominant role. An observation is a real object containing
both methods and data. One observation is instantiated for every
sensor-landmark pair, regardless of the sensor having actually observed the
landmark or not, and has the same lifetime as the associated landmark. 
The methods it contains are the conventional projection and back-projection
models (that depend on the associated sensor and landmark models),
while the stored data consist of results and intermediary variables
such as Jacobian matrices, predicted and observed appearances, innovations,
event counters and others, that allow to greatly simplify and optimize
computations.%

\paragraph{Managers.}
In order to achieve full genericity \textit{wrt} landmark types, in particular
to allow the concurrent use of different landmark types for one sensor, two
different manager objects are added: \emph{data manager} and \emph{map
  manager}. Their implementations define a given management strategy, while
their instantiations are dedicated to a certain landmark type.
The data manager processes the sensors raw data, providing observations of the
external landmarks. For this purpose, it exploits some raw data processors
(for feature detection and tracking),
and decides which observations are to be corrected and in
which order, according to the quantity of information they bring and their
quality. For example it can apply an active search strategy and try to
eliminate outliers as described in section \ref{sec-updatestrategy}.
The map manager keeps the map clean, with relevant information, 
and at a manageable size, by removing landmarks according to their quality and the
given policy ({\em e.g.} visual odometry where landmarks are discarded once they
are not observed, or multimap slam where maps are ``closed'' according to given
criteria).
 These managers communicate together: for example, the data manager may 
ask the map manager if there is enough space in the map to start a new
 initialization process, and to allocate the appropriate space for the new
 landmark.

\section{Inertial/visual SLAM within \rtslam\label{sec-technique}}

\subsection{Active search and one-point RANSAC\label{sec-updatestrategy}}

The strategy currently implemented in \rtslam's data manager to deal
with observations is an astute combination of
\emph{active search} \cite{Dav_07} and outliers rejection using
\emph{one-point RANSAC} \cite{civera_onepointransac}.

The goal of active search is to minimize the quantity of raw data processing by
constraining the search in the area where the landmarks are likely to be
found. Observations outside of this 3$\sigma$ observation uncertainty ellipse
would be anyway considered incompatible with the filter and ignored by the
\emph{gating} process. In addition active search gives the possibility to decide
anytime to stop matching and updating landmarks with the current available data,
thus enabling \emph{hard real-time} constraints. We extended the active search
strategy to landmark initialization: each sensor strives to maintain features in
its whole field of view using a randomly moving grid of fixed size, and feature
detection is limited to empty cells of the grid. Furthermore the good
repartition of features in the field of view ensures a better observability of
the motions.

Outliers can come from matching errors in raw data or mobile objects in the
scene. Gating is not always discriminative enough to eliminate them,
particularly right after the prediction step when the observation uncertainty
ellipses can be quite large -- unfortunately at this time the filter is very
sensitive to faulty corrections because it can mistakenly make all the following
observations incompatible. To prevent faulty observations, outliers are rejected
using a one-point RANSAC process. It is a modification of RANSAC, that uses the
Kalman filter to obtain a whole model with less points than otherwise needed,
and provides a set of \emph{strongly compatible} observations that are then
readily corrected. Contrary to \cite{civera_onepointransac} where data
association is assumed given when applying the algorithm, we do the data
association along with the one-point RANSAC process: this allows to look for
features in the very small strongly compatible area rather than the whole
observation uncertainty ellipse, and to save additional time for raw data
processing.

\subsection{Landmark parametrizations and reparametrization\label{sec-param}}

In order to solve the problem of adding to the EKF a point with unknown distance
and whose uncertainty cannot be represented by a Gaussian distribution, point
landmarks parametrizations and initialization strategies for monocular EKF-SLAM
have been well studied \cite{Dav_03} \cite{lemaire_delayed}
\cite{Montiel_06}. The solutions now widely accepted are undelayed
initialization techniques with \emph{inverse depth} parametrization. Anchored
Homogeneous Point \cite{sola_ahp} parametrization is currently used in \rtslam.

The drawback of inverse depth parametrizations is that they describe a landmark
by at least 6
variables in the stochastic map, compared to only 3 for an Euclidean point
$\left(x\ y\ z\right)^T$.  Memory and temporal complexity being quadratic with
the map size for EKF, there is a factor of 4 to save in time and memory by
\emph{reparametrizing} landmarks that have converged enough \cite{Civera07}. The
map manager uses the linearity criterion proposed in \cite{Montiel_06} to
control this process.

\subsection{Motion prediction \label{sec-prediction}}

The easiest solution for EKF-SLAM is to use a robot kinematic model
to do the filter prediction, such as a constant velocity model:
$$\mathcal R = \left({\bf p}\ {\bf q}\ {\bf v}\ {\bf w}\right)^T$$
where  $\bf p$ and $\bf q$ are respectively the position and quaternion orientation
of the robot, and $\bf v$ and $\bf w$ are its linear and angular
velocities.

The advantage of such a model is that it does not require complicated hardware
setup (only a simple camera), but its strong limitation is that the scale factor
is not observable. A second camera with a known baseline can provide a proper
scale factor, but one can also use a proprioceptive sensor for the prediction
step. Furthermore it usually provides a far better prediction with smaller
uncertainties than a simple kinematic model, which brings several benefits:
\begin{enumerate}
\item search areas for matching are smaller, so processing is
  faster,
\item linearization points are more accurate, so SLAM precision is better,
 or one can reduce the framerate to decrease CPU load for equivalent quality,
\item it allows to withstand high motion dynamics.
\end{enumerate}

In the case of an Inertial Measurement Unit (IMU), the robot state is then:
$$\mathcal R = \left({\bf p}\ {\bf q}\ {\bf v}\ {\bf a_b} \ {\bf
w_b}\ {\bf g}\right)^T$$
where $\bf a_b$ and $\bf w_b$ are the accelerometers and gyrometers biases, and 
$\bf g$ the 3D gravity vector. Indeed it is better for linearity reasons to
estimate the direction of $\bf g$ rather than constraining the robot
orientation to enforce $\bf g$ to be vertical.

A special care has to be taken for the conversion of the noise
from continous time (provided by the manufacturer in the sensor's datasheet) to discrete time. As the perturbation is continuous white noise, the variance of the discrete noise grows linearly with the integration period.

\subsection{Image processing\label{sec-image-proc}}

\paragraph{Point extraction.}

Point extraction is based on Harris detector with several optimizations.  Some
of them are approximations: a minimal derivative mask $[-1,0,1]$ is used, as well as a
square and constant convolution mask, in order to minimize operations.  This
allows the use of \emph{integral images} \cite{Viola_integral} to efficiently
compute the convolutions.  Additional optimizations are related to active search
(section \ref{sec-updatestrategy}): only one new feature is searched in a small region of
interest, which eliminates the costly steps of thresholding and sub-maxima
suppression.

\paragraph{Point matching.}

Point matching is based on Zero-mean Normalized Cross Correlation (ZNCC), 
also with several optimizations. Integral images are used to compute means and
variances, and a hierarchical search is made (two searches at half and full
resolution are sufficient). We also implemented \emph{bounded partial
  correlation} \cite{DiStefano05} in order to interrupt the correlation score
computation when there is no more hope to obtain a better score than the
threshold or the best one up to now. To be robust to viewpoint changes and to
track landmarks longer, tracking is made by comparing the initial appearance of
the landmark with its current predicted appearance \cite{Dav_07}.

\section{Results\label{sec-results}}

\begin{figure}[ht]
	\centering
	\includegraphics[height=2.65cm]{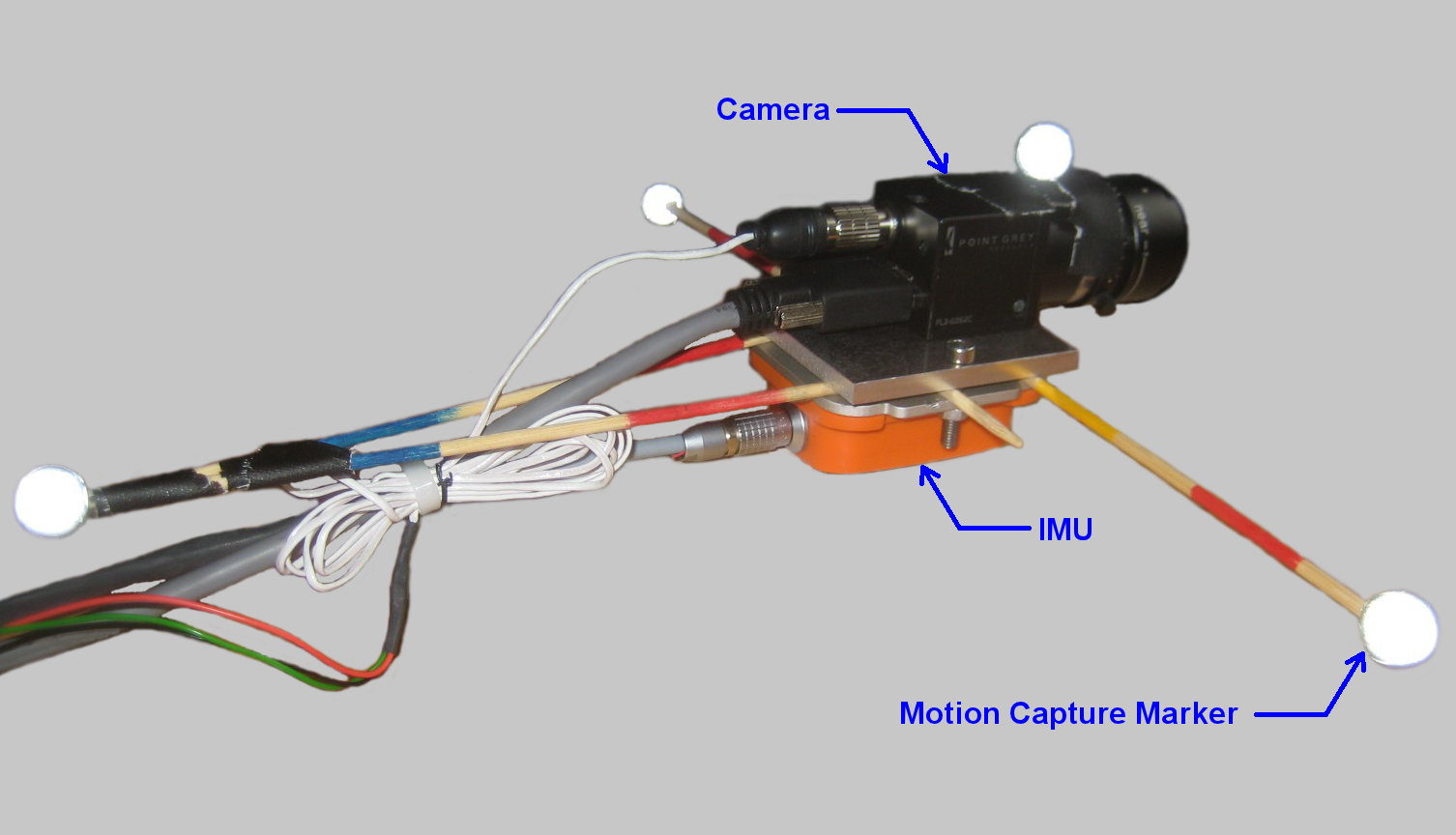}
	\includegraphics[height=2.65cm]{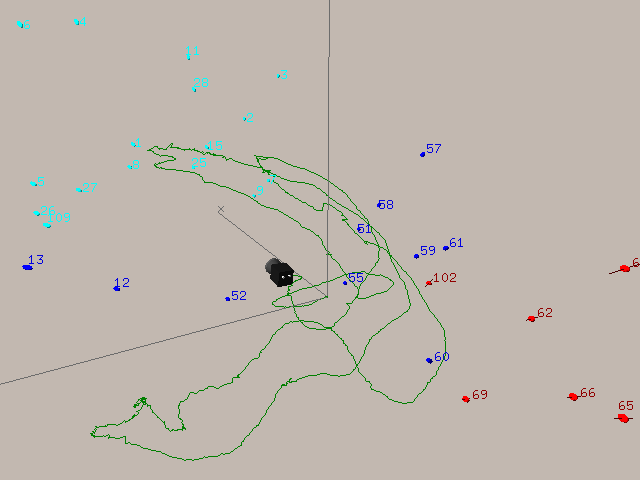}
	\includegraphics[height=2.65cm]{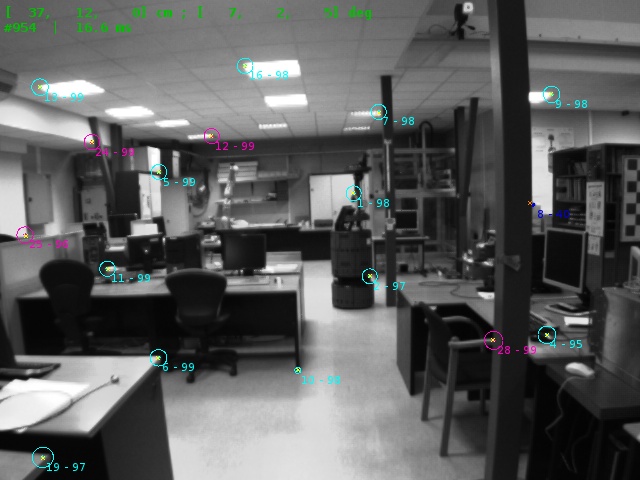}
	\caption{The experimental setup composed of a Flea2 camera and an XSens MTi IMU,
 and screen captures of the 3D and 2D display of \rtslam.}
	\label{res_setup}
\end{figure}

Fig. \ref{res_setup} shows the hardware setup that has been used for the
experiments. It is composed of a firewire camera rigidly tied to an IMU, on
which four motion capture markers used to measure the ground truth are fixed.

Two different sequences are used, referred to as \emph{low dynamic} and 
\emph{high dynamic} sequences. Both were acquired indoor with artificial
lights only, with an image framerate of 50\,Hz synchronized to the inertial data rate of 100\,Hz.
The motion capture markers are localized with a precision of approximately
1\,mm, so the ground truth has a precision of $\sigma_{xyz}=1\,$mm in position
and $\sigma_{wpr}=0.57^{\circ}$ in angle (baseline of 20\,cm).

Movies illustrating a run of every experiment are provided
\ifx\review\reviewfalse
at:\\\texttt{http://rtslam.openrobots.org/Material/ICVS2011}.
\else
along with the article submission.
\fi

\subsection{Constant velocity model}

\begin{figure}[htbf!]
	\includegraphics[width=\linewidth]{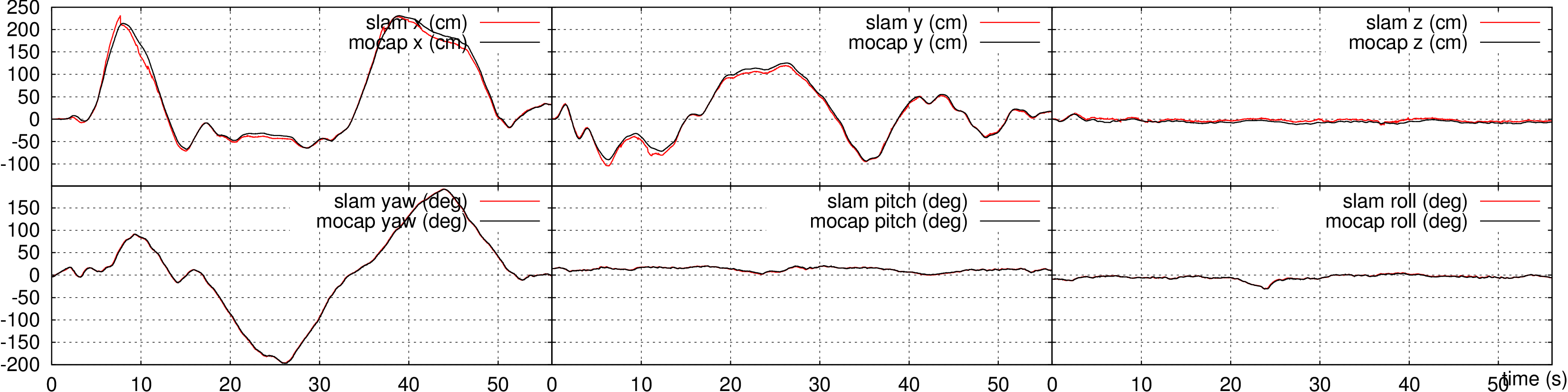}
	\caption{Illustration of low dynamic trajectory, constant velocity SLAM 
(with scale
          factor of 2.05 manually corrected). Estimated trajectory parameters
          and ground truth, as a function of time (in seconds).}
	\label{res_ld_traj}
\end{figure}

\begin{figure}[htbf!]
	\includegraphics[width=\linewidth]{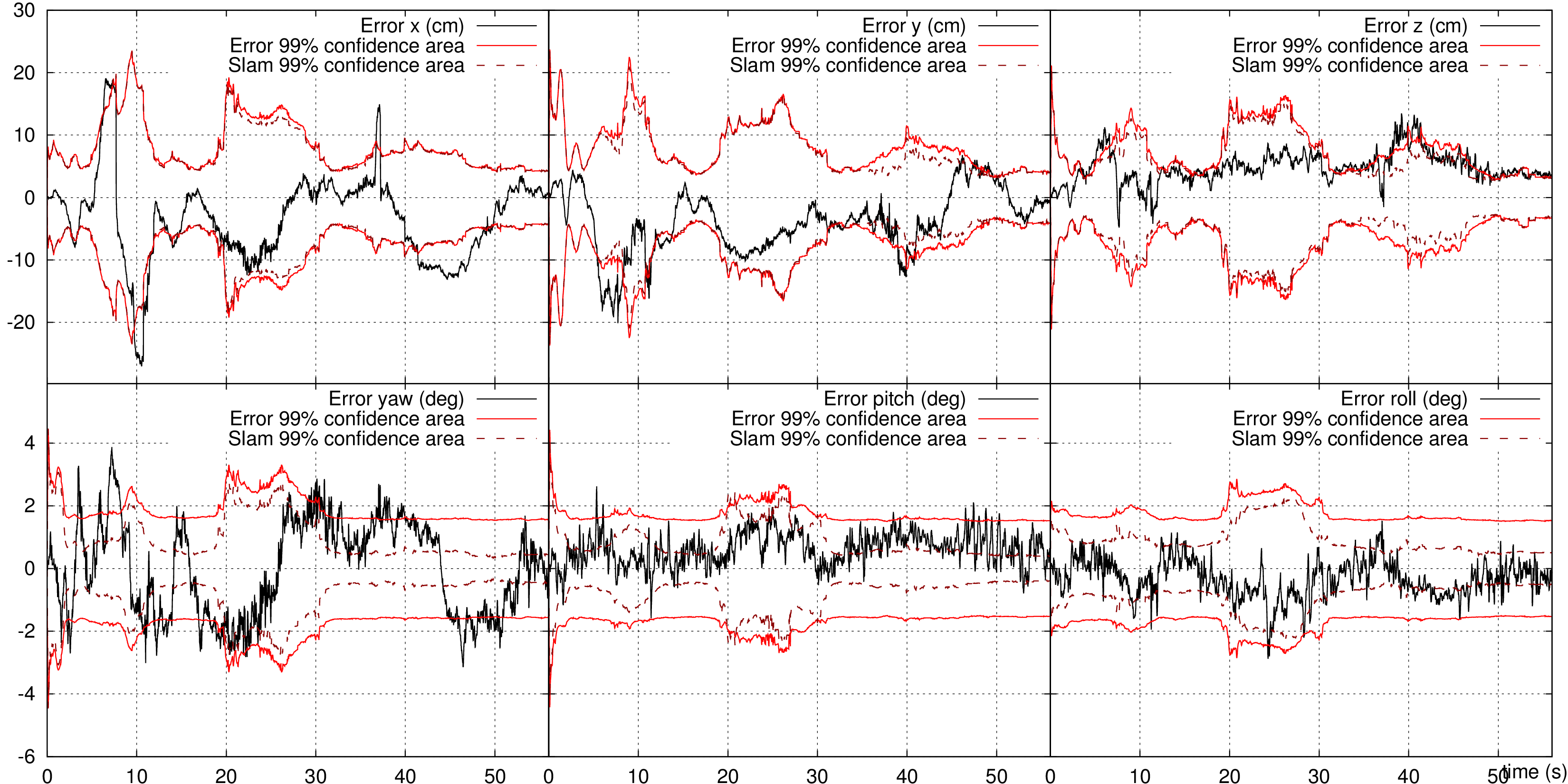}
	\caption{Errors of the estimated parameters of the trajectory shown Fig.
          \ref{res_ld_traj}. The 99\% confidence area corresponds to $2.57\sigma$
          bounds. The SLAM 99\% confidence area, that does not include ground
          truth uncertainty, is also shown.}
	\label{res_ld_cv_error}
\end{figure}

The inertial data is here not used, and the prediction is made according to a
constant velocity model. Fig. \ref{res_ld_traj} presents the estimated
trajectory and the ground truth for the \emph{low dynamic} sequence, and
Fig. \ref{res_ld_cv_error} shows the absolute errors for the same run.

Besides the global scale factor which is manually corrected, the camera
trajectory is properly estimated. The movie shows that loops are correctly
closed, even after a full revolution.

The position error raises up to 10\,cm after quick rotations of the camera: this
is due to a slight \emph{drift} of the scale factor caused by the integration of
newer landmarks (when closing loops, the scale factor is restimated closer to
its initial value). Also, uncertainty ellipses remain relatively large throughout
the sequence: these issues are solved by the use of an IMU to predict the
motions.

\subsection{Inertial/visual SLAM}

\begin{figure}[htbf!]
	\includegraphics[width=\linewidth]{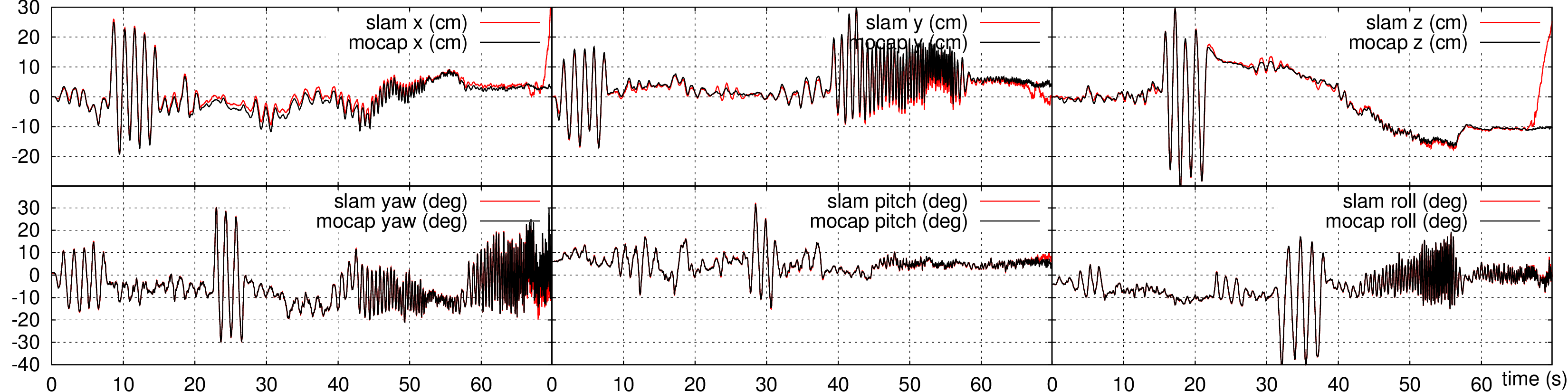}
	\caption{Illustration of high dynamic trajectory, inertial/visual SLAM.
 Estimated trajectory
          parameters and ground truth, as a function of time (in seconds).}
	\label{res_hd_traj}
\end{figure}

\begin{figure}[htbf!]
	\includegraphics[width=\linewidth]{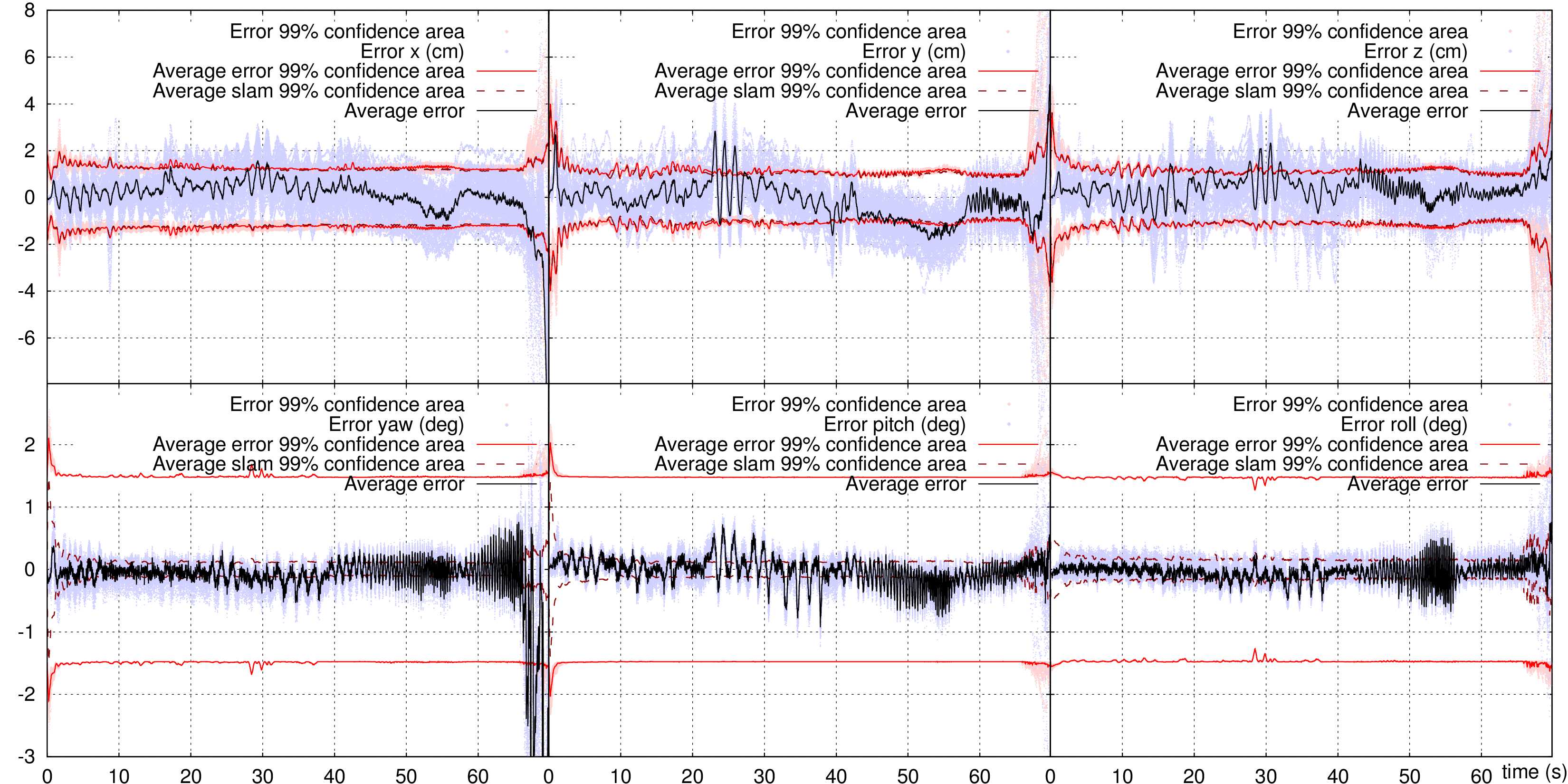}
	\caption{Errors of inertial/visual SLAM estimated over 100 runs on the
          dynamic sequence, as a function of time (in seconds) -- the difference between 
          each run is due to the randomized landmark selection,
          see section \ref{sec-updatestrategy}.
          All the runs remain in the
          neighborhood of the 99\% confidence area. The angular ground truth
          uncertainty is not precisely known: its theoretical value is
          both overestimated and predominant over SLAM precision in such a
          small area.}
	\label{res_hd_error}
\end{figure}

\begin{figure}[htbf!]
	\includegraphics[width=\linewidth]{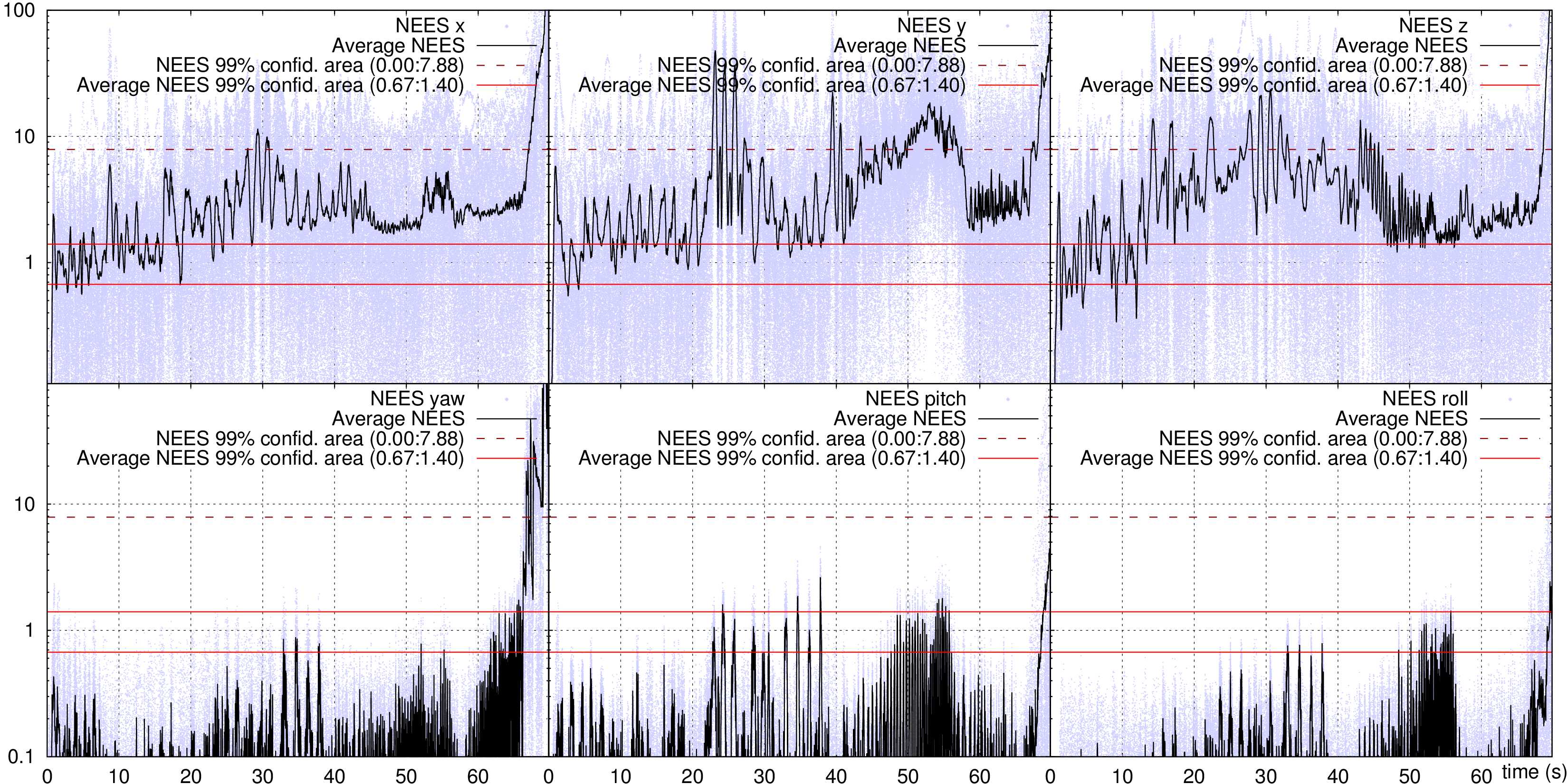}
	\caption{Inertial/visual SLAM NEES \cite{Bailey06} over 100 runs on the dynamic sequence.
          The average NEES is quickly out of the corresponding 99\%
          confidence area, but as the 100 runs were made on the same sequence
          they are not independant and the average NEES should rather be
          compared to the simple NEES confidence area. The filter appears to be
          very conservative with angles but this is due to the overestimated
          ground truth uncertainty as explained in Fig. \ref{res_hd_error}.}
	\label{res_hd_nees}
\end{figure}

Fig. \ref{res_hd_traj} shows the trajectory estimated with the \emph{high
  dynamic} sequence, and Fig. \ref{res_hd_error} and \ref{res_hd_nees} show the
behavior of inertial SLAM. 
Here, all of the 6 degrees of freedom are successively excited,
then $y$ and $yaw$ are excited with extreme dynamics:
the yaw angular velocity goes up to 400\,deg/s,
the rate of change of angular velocity exceeds 3,000\,deg/s$^2$,
and accelerations reach 3\,$g$.
It is interesting to note that the time when SLAM diverges (around $t=65$\,s)
corresponds to motions for which the angular velocity exceeds the limit of the
IMU (300\,deg/s) and where its output is saturated.

The IMU now allows to observe the scale factor, and at the same time reduces the
observation uncertainty ellipses and thus eases the active search procedure.
Conversely, the divergence of the SLAM process at the end of the sequence
illustrates what happens when vision stops stabilizing the IMU: it quickly
diverges because the biases and the gravity cannot be estimated anymore.

\section{Outlook\label{sec-conclusion}}

We have presented a complete SLAM architecture whose genericity and performance
make it both a useful experimentation tool and efficient solution for robot
localization. The following extensions are currently being developed: using
a second camera to improve landmarks observability, a multimap
approach to cope with large scale trajectories and multirobot
collaborative localization, and the use of line landmarks complementarily to
points to provide more meaningful maps and to ease map matching for loop
closure. 

The architecture of \rtslam\ allows to easily integrate such developments, and
also to consider additional motion sensors: to increase the robustness of the
system, it is indeed essential to consider the various sensors usually found
on board a robot (odometry, gyrometers, GPS). Eventually, it would be
interesting to make \rtslam\ completely generic \emph{wrt} the estimation engine
as well, in order to be able to use global optimization techniques in place of
filtering.

\bibliographystyle{plain}
\bibliography{biblio.bib}

\end{document}